\documentclass[conference]{IEEEtran}
\IEEEoverridecommandlockouts
\usepackage{cite}
\usepackage{amsmath,amssymb,amsfonts}
\usepackage{algorithmic}
\usepackage{graphicx}
\usepackage{textcomp}
\usepackage{xcolor}
\usepackage{algorithm}
\usepackage{algorithmic}
\usepackage{epsfig}
\usepackage{graphicx}
\usepackage{amsmath}
\usepackage{multirow}
\usepackage{bbding}
\usepackage{color,xcolor}
\usepackage{colortbl}
\usepackage{subcaption}
\usepackage{booktabs}
\usepackage{amssymb,amsfonts}
\usepackage{bm}
\usepackage{amsthm}
\usepackage{color}
\usepackage{lineno}
\usepackage{pifont}
\usepackage[colorlinks,hidelinks]{hyperref}
\definecolor{airforceblue}{rgb}{0.36, 0.54, 0.66}
\definecolor{babyblue}{rgb}{0.54, 0.81, 0.94}
\definecolor{babyblueeyes}{rgb}{0.63, 0.79, 0.95}

\hypersetup{
    colorlinks=true,
    linkcolor=black,
    filecolor=magenta,      
    urlcolor=purple
    }

\def\BibTeX{{\rm B\kern-.05em{\sc i\kern-.025em b}\kern-.08em
    T\kern-.1667em\lower.7ex\hbox{E}\kern-.125emX}}
\begin{document}

\title{EmoHead: Emotional Talking Head via \\Manipulating Semantic Expression Parameters}


\author{Xuli Shen$^{1,2}$ \ \  Hua Cai$^{2}$ \ \  Dingding Yu$^{2}$ \ \ Weilin Shen$^{2}$ \ \  Qing Xu$^{2}$  \ \   Xiangyang Xue$^{1 \star}$  \\ $^{1}$Fudan University  \ \ \ \ \ \ \ \ \ \ \ \ \ \ \ \ \ \ $^{2}$UniDT  \\ \href{https://bisno.github.io/Emohead}{\textit{ https://bisno.github.io/Emohead}}
         }

\maketitle

\begin{abstract}
Generating emotion-specific talking head videos from audio input is an important and complex challenge for human-machine interaction. However, emotion is  highly abstract concept with ambiguous boundaries, and it necessitates disentangled expression parameters to generate emotionally expressive talking head videos. In this work, we present EmoHead to synthesize talking head videos via semantic expression parameters. To predict expression parameter for arbitrary audio input, we apply an audio-expression module that can be specified by an emotion tag. This module aims to enhance correlation from audio input across various emotions. Furthermore, we leverage  pre-trained hyperplane to refine facial movements by probing along the vertical direction. Finally, the refined expression parameters regularize neural radiance fields and facilitate the  emotion-consistent generation of talking head videos. Experimental results demonstrate that semantic expression parameters lead to better reconstruction quality and controllability.
\end{abstract}

\begin{IEEEkeywords}
emotional talking head synthesis, video manipulation, multi-modality alignment 
\end{IEEEkeywords}

\section{Introduction}
\let\thefootnote\relax\footnotetext{$^\star$Corresponding author.}Talking head synthesis is the process of reanimating a target person to align with  lip movement and head poses executed by input audio \cite{prajwal2020lip,guo2021ad,song2022everybody, lipsHD23,Tan_2024}. It involves the seamless integration of the audio source and target avatar, resulting in a realistic and synchronized video output by harnessing modern generative reconstruction methods \cite{goodfellow2014generative,mildenhall2020nerf}.



Emotional talking head synthesis, on the other hand, aims to generate talking avatar videos with vivid expressions for specific emotions \cite{wang2022one,liang2022expressive,Gan_2023_ICCV}. Benefiting from the facial morphable model \cite{blanz1999morphable,guo2020towards}, there are end-to-end methods that focus on audio-driven \cite{liang2022expressive} or video-driven \cite{Gan_2023_ICCV} pipelines that empower with facial expression editing capabilities. \cite{tan2025edtalk}.

\begin{figure}
 \includegraphics[width=0.99\linewidth]{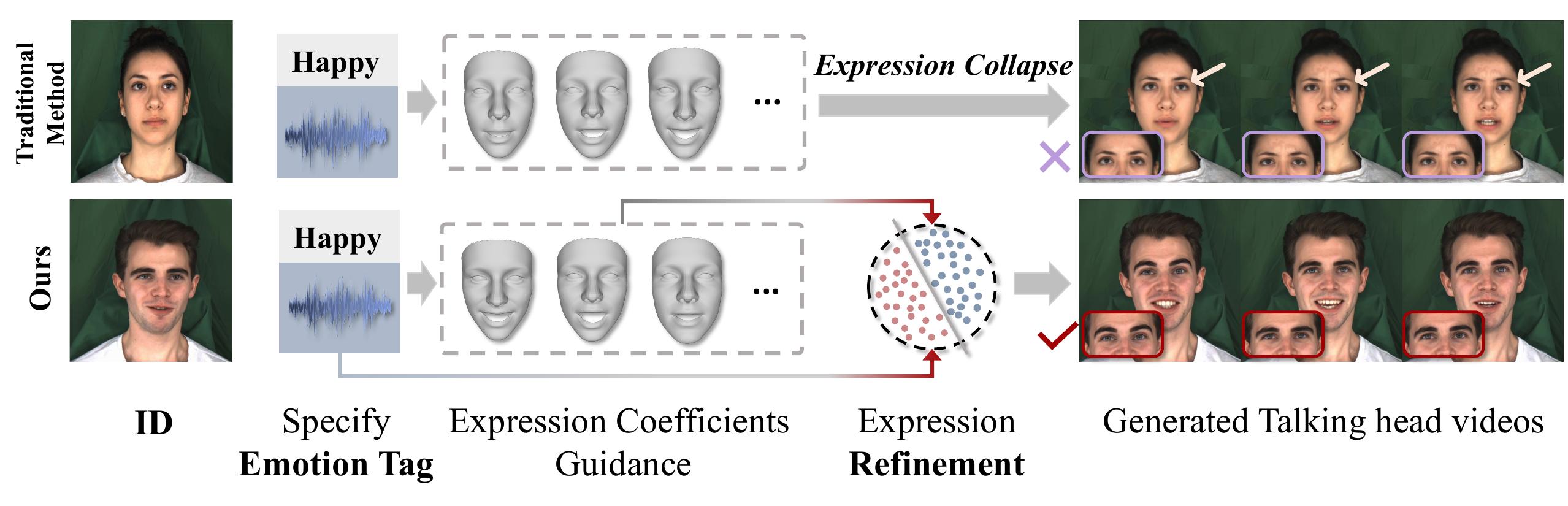}
 \centering
  \caption{Different from previous work, the proposed method applies emotion-specific hyperplanes to eliminate ``expression collapse'' phenomenon and generate target emotional videos.  }
\label{fig:teaser}
\vspace{-0.15in}
\end{figure}

Nevertheless, these expression parameters are often entangled, which leads to the ``\textbf{expression collapse}" phenomenon in reconstructed results, as depicted by the angry eye region for the target happy emotion in Fig.~\ref{fig:teaser}.  We hypothesize that if the expression parameters are sufficiently refined, this phenomenon will be eliminated and emotional consistency will be improved, as seen in the smiling eye region of Fig.~\ref{fig:teaser}. In this work, we  propose an target emotion expression refinement method to enhance the overall authenticity and effectiveness of emotional talking head synthesis. By accurately representing and refining facial movements, the framework enables more engaging and emotionally resonant experiences. 

Specifically, we present an audio-expression module that regresses low-dimensional expression parameters based on audio input from various sources and specific emotion categories. To better synchronize emotional talking videos,  we first obtain the embedded audio feature and use timestamp-aligned audio speech recognition alongside pretrained audio-emotion encoder and text encoder to extract emotion features in audio and text features, respectively. We then  propose an emotional cross-modality fusion mechanism, named Audio Expression Alignment, to eliminate noise emotions and preserve strong correlations. Next, we train  emotion-specific hyperplanes to intricately refine and control facial movements through the low-dimensional expression parameters. Next, we leverage the predicted expression parameters and construct emotion-conditional implicit function for talking head reconstruction using volume rendering, and achieve  high-fidelity synthesis with better controllability. Our key contributions are:
\begin{itemize}
\item  We present an audio-expression module for mapping audio features to low-dimensional expression parameters. We propose Audio Expression Alignment, a cross-modality attention mechanism that enhances the correlation between audio features and  emotion target.
\item We apply an expression refinement method to enhance the emotion consistency of  expression parameters and  manipulate  emotion representation in  the talking stage.
\item We leverage refined facial expression parameters for portrait rendering, resulting in better reconstruction quality and  controllability of the target emotion.
\end{itemize}

\begin{figure*}[t]
\begin{center}
\includegraphics[width=0.8\linewidth]{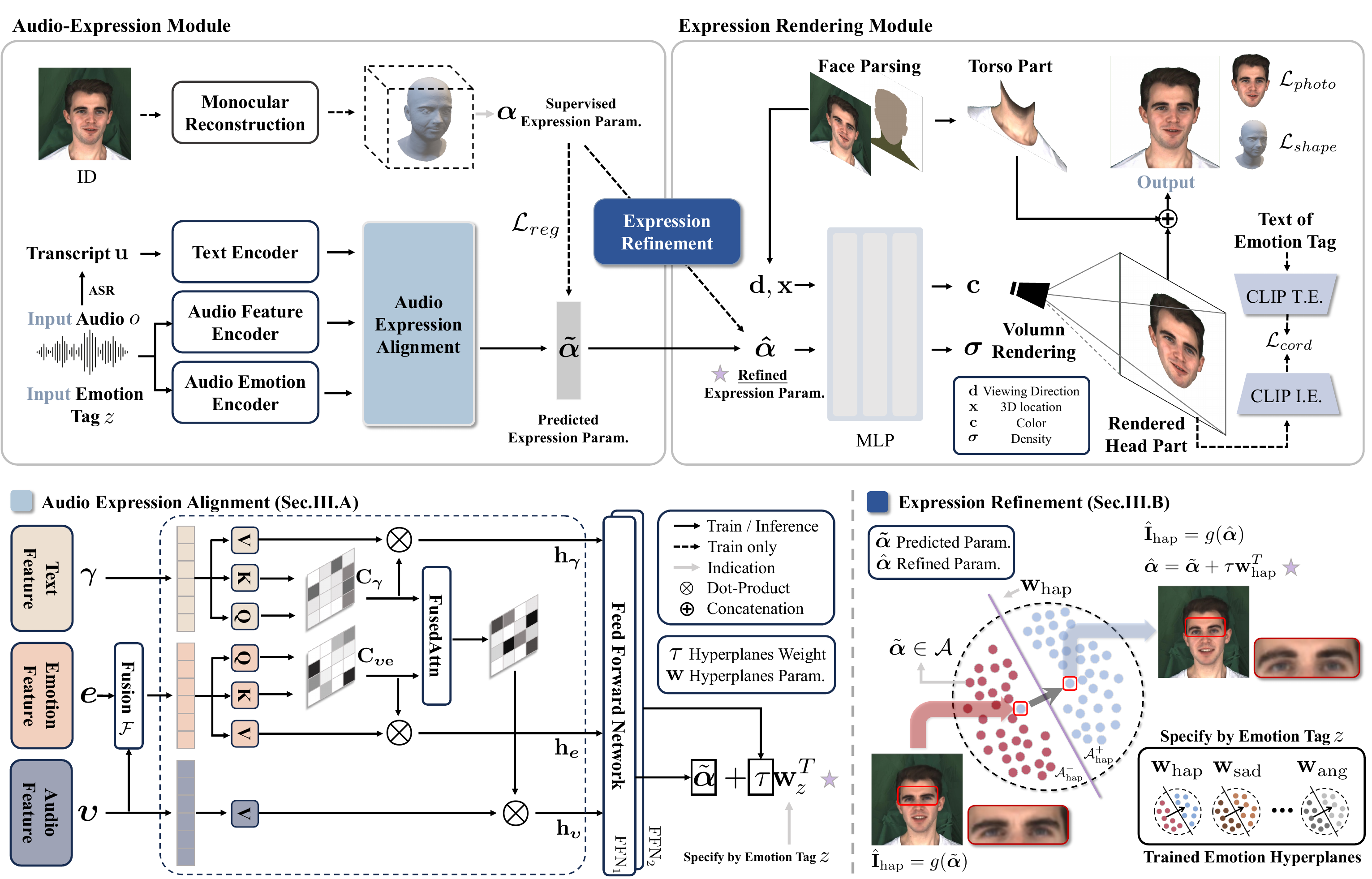}
\end{center}
\caption{The proposed framework of EmoHead. }
\label{fig:framwork}
\vspace{-0.15in}
\end{figure*}

\section{Related Work}

\subsection{Audio-Driven Talking Head Generation}
Existing audio-driven talking head generation  can be divided into identification-specific \cite{guo2021ad,song2022everybody, lipsHD23,Tan_2024} and identification-agnostic \cite{ ma2023styletalk,guo2024liveportrait} fields. 
Identification-specific audio-driven talking head generation aims to synthesize videos of specifically trained target individuals. Recently, Style2Talker \cite{Tan_2024} learns personalized speaker-specific diffusion models using 3DMM coefficients and then generates videos via CLIP \cite{clip_short} guidance. In the field of identification-agnostic methods, talking head videos are generated in a one-shot setting, building a universal model applicable across different individuals. Specifically, SadTalker \cite{zhang2022sadtalker} learns decoupled and realistic motion coefficients from audio inputs. LivePortrait \cite{guo2024liveportrait} drives arbitrary talking faces with audio while allowing free pose control by learning pose motions from another  video source.

\subsection{Emotional Talking Head Generation}
Emotional talking head generation requires both emotional manipulation and high-fidelity generation method. In recent years, \cite{seeskimezemotface,wang2022one, liang2022expressive,ji2022eamm}  has focused  on generating one-shot talking head animations with emotion control by providing an identity source image and an emotion source video. EAMM \cite{ji2022eamm} leverages  emotion discriminative loss for emotion talking head videos. EDTalk \cite{tan2024edtalk} aims to decompose the facial dynamics to enhance emotion representation. EAT \cite{Gan_2023_ICCV} focuses on controlling the emotion via  prompt. Nevertheless, few efforts have attempted to explore precise emotion representation for emotional talking head synthesis in cases where different emotions couple together.

\section{Method}

In this section, we present emotional talking head synthesis framework EmoHead, illustrated in Fig.~\ref{fig:framwork}. First, we demonstrate the audio-expression module, which  predicts facial expression parameters from multi-modality features. Next, we introduce a novel expression parameters refinement via  trained hyperplane that classifies the target emotion. Then, we elaborate on the expression rendering module, which uses the expression parameters to synthesize 2D talking head videos via neural radiance field. Finally, we discuss  key training details.

\noindent \textbf{Task Formulation}. We aim to synthesize  talking face video given specific emotion tag. The input audio state  $\mathbf{s}$  is a pair $ (o, z) $ where $o$ is a fixed-length audio and $z $ is the tag of required emotion state. The goal of the task is to learn a function $\mathcal{G}: \mathbf{s} \rightarrow  \boldsymbol{\alpha} \rightarrow  \mathcal{I}$ , where $\mathcal{I}$ is the  set of output frames and  $\boldsymbol{\alpha}\in \mathbb{R}^{m}$ refers to expression parameter. 

\subsection{Audio-Expression Module
 \label{sec:4} }

Given the audio input frame $i$, we obtain the audio features $\tilde{\boldsymbol{\upsilon}}_i$, and emotional features  $\tilde{\textbf{e}}_i$ present in the audio $o_i$ using HuBERT-base as audio feature encoder \cite{soft-vc-2022} and Emotion2vec-base \cite{ma2023emotion2vec} as audio emotion encoder. Next, we utilize audio speech recognition (ASR \cite{gao2023funasr}) to provide  transcript $\textbf{u}$, which is timestamp-aligned by audio (see Appendix.C). Then, we leverage the large language model LLaMa2-7b \cite{touvron2023llama} to encode text features $\tilde{\boldsymbol{\gamma}}_i$. We apply three different trainable linear mappings $\textbf{E}$ to project all the above features into same dimension $d$:
\begin{equation}
 \boldsymbol{\upsilon}_i =    \tilde{\boldsymbol{\upsilon}}_i \textbf{E}_{1}; \  \textbf{e}_i =    \tilde{\textbf{e}}_i \textbf{E}_{2}  ; \   \boldsymbol{\gamma}_i =    \tilde{\boldsymbol{\gamma}}_i \textbf{E}_{3}.
\end{equation}


\noindent \textbf{Audio Expression Alignment}. To better preserve lip-synchronization with various audio emotions, we leverage cross-modality feature alignment to predict expression parameters. Specifically, we first fuse audio feature and emotion feature and compute the correlation matrix as
\begin{equation}
\textbf{C}_{\boldsymbol{\upsilon}\textbf{e}} = \mathcal{F}(\boldsymbol{\upsilon},\textbf{e}) \textbf{W}_{\text{q},1}\textbf{W}_{\text{k},1}^T  \mathcal{F}(\boldsymbol{\upsilon},\textbf{e}) ^T ,
\end{equation}
where $\textbf{W}_{\text{q},1}, \textbf{W}_{\text{k},1}\in \mathbf{R}^{d \times d_h}$ are the projection matrices. $\mathcal{F}$ denotes the fusion function from the neighboring frames, e.g., $ \mathcal{F}(\boldsymbol{\upsilon},\textbf{e})  = \sum_{i-n}^{i}(\boldsymbol{\upsilon}_i + \textbf{e}_i)$, and $n$ is empirically set as 5. 
Next, we compute the correlation matrix of the text feature as:
\begin{equation}
\textbf{C}_{\boldsymbol{\gamma}} = \boldsymbol{\gamma}_i \textbf{W}_{\text{q},2}\textbf{W}_{\text{k},2}^T  \boldsymbol{\gamma}_i^T ,
\end{equation}
where $\textbf{W}_{\text{q},2}, \textbf{W}_{\text{k},2}\in \mathbf{R}^{d \times d_h}$. Then, to better align audio and emotion feature, we fuse the two correlation matrices and obtain the attention formula as follows (denoted as ``FusedAttn''):
\begin{align}
\textbf{h}_{\boldsymbol{\upsilon}} & = \text{FusedAttn}(\boldsymbol{\upsilon},\textbf{e},\boldsymbol{\gamma}) \\
&= \text{Softmax}(\frac{\textbf{C}_{\boldsymbol{\upsilon}\textbf{e}}+\textbf{C}_{\boldsymbol{\gamma}}}{\sqrt{d_h}})\boldsymbol{\upsilon}\textbf{W}_{\text{v},1},
\end{align}
where $\textbf{W}_\text{v,1}\in \mathbf{R}^{d \times d}$ and $\textbf{h}_{\boldsymbol{\upsilon}}$ denotes the hidden state of the audio feature. Finally, we consider the fused emotion and text feature important enough to be fully learned, so we also pass and update them between separately as follows:
\begin{align}
\textbf{h}_{\boldsymbol{e}}
&= \text{Softmax}(\frac{\textbf{C}_{\boldsymbol{\upsilon}\textbf{e}}}{\sqrt{d_h}})\mathcal{F}(\boldsymbol{\upsilon},\textbf{e})\textbf{W}_{\text{v},2} \\
\textbf{h}_{\boldsymbol{\gamma}} &=  \text{Softmax}(\frac{\textbf{C}_{\boldsymbol{\gamma}}}{\sqrt{d_h}})\boldsymbol{\gamma}\textbf{W}_{\text{v},3} ,
\end{align}
where $\textbf{W}_\text{v,2},\textbf{W}_\text{v,3} \in \mathbf{R}^{d \times d}$ and $\textbf{h}_{\boldsymbol{e}}$ and $\textbf{h}_{\boldsymbol{\gamma}}$  denote the
hidden states of the emotion and the text features, respectively. At last,  we have the concatenated feature $[\textbf{h}_{\boldsymbol{\upsilon}};\textbf{h}_{\textbf{e}};\textbf{h}_{\boldsymbol{\gamma}}]$, which is sent to  feed-forward networks to predict expression parameter $ \tilde{\boldsymbol{\alpha}}$. Note that $ \tilde{\boldsymbol{\alpha}}$   is supervised by the $\boldsymbol{\alpha}$ from 3DMM expression coefficients \cite{guo2020towards}.



\subsection{Expression Refinement} \label{sec:emosec3} 
There remains a non-trivial challenge that the predicted  $ \tilde{\boldsymbol{\alpha}} $ can not precisely generate target emotion. The rendered frame with  ``happy'' emotion  displays expression collapse of twinkling eyebrows in the bottom-right of Fig.~\ref{fig:framwork}. A simple method is to  pre-set certain dimension of expression parameter $ \tilde{\boldsymbol{\alpha}} $ to edit the emotion, for example setting $\text{dim}_6 = -2.1$ (depicted in Fig.1 of Appendix.A) to alleviate coupled emotion representation, but not all dimensions of expression parameters are relevant to  specific emotion individually.


To this end, we employ emotion specific hyperplanes to explore the entangled expression parameter space, depicted in the bottom right part of Fig.~\ref{fig:framwork}. These hyperplanes, pre-trained by frames of  emotion specified talking head videos, enable the model to determine the boundary of emotion concept and refine parameter $ \tilde{\boldsymbol{\alpha}} $ to \textit{\textbf{semantic expression parameters}}. Specifically, the  hyperplane is set as classifier in the expression parameter space $ \mathcal{A} $. We define $ \{ y^{+}, y^{-} \}$ to label the target emotion parameters $ \mathcal{A}^{+} $ and  parameters for other emotions $ \mathcal{A}^{-} $. The  hyperplane parameter is defined as
$ \mathbf{w}_z \in \mathbb{R}^{m} $  with respect to emotion $z$. After obtaining the trained hyperplanes (see Appendix.B for training details), the  frames can be generated and  refined via predicted expression parameter along the normal vector direction:
\begin{linenomath}
\begin{align}
 \tilde{\boldsymbol{\alpha}}  = \text{FFN}_1([\textbf{h}_{\boldsymbol{\upsilon}};\textbf{h}_{\textbf{e}} & ;\textbf{h}_{\boldsymbol{\gamma}}]) , \tau = \text{FFN}_2([\textbf{h}_{\boldsymbol{\upsilon}};\textbf{h}_{\textbf{e}};\textbf{h}_{\boldsymbol{\gamma}}]) \\
& \hat{\boldsymbol{\alpha}} = \tilde{\boldsymbol{\alpha}} + \tau \mathbf{w}_z^{T}  \\
& \hat{\mathbf{I}} = g(\hat{\boldsymbol{\alpha}}) ,
\label{eq:intecthp}
\end{align}
\end{linenomath}
where FFN refers to  feed-forward network, $ \tau $ determines a learnable weight of manipulation, $ \tilde{\boldsymbol{\alpha}} \in \mathcal{A} $ refers to the predicted expression parameter and $ g $ is the expression rendering module to generate frame $\hat{\mathbf{I}}$ and will be introduced in next section. Therefore, the emotion consistency of rendered talking head is enhanced by changing the expression parameter along the direction of normal vector, resulting in ``happy'' emotion with relaxed eyebrows   in bottom right part of Fig.~\ref{fig:framwork}. 





\subsection{Expression Rendering Module
}
\noindent \textbf{Neural Radiance Fields for EmoHead}. 
Different from  previous audio-driven reconstruction methods \cite{guo2021ad,ye2024real3d}, we present an expression-conditioned neural radiance field. We apply  implicit function with inputs of a viewing direction $\mathbf{d}$, a 3D location $\mathbf{x}$, and refined expression parameters $\hat{\boldsymbol{\alpha}}$. This implicit function  $\theta$ is realized by multi-layer perceptron (MLP). By combining the input vectors $(\hat{\boldsymbol{\alpha}}, \mathbf{d}, \mathbf{x})$, the MLP estimates  colors $\mathbf{c}$ and densities $\boldsymbol{\sigma}$ of  rays. This implicit function is  formulated as  
$\theta: (\hat{\boldsymbol{\alpha}}, \mathbf{d}, \mathbf{x}) \rightarrow (\mathbf{c}, \boldsymbol{\sigma}).
\label{eq:nerf}$
Similar to the rendering process in NeRF \cite{mildenhall2020nerf}, the expected color $\mathcal{C}$ of a camera ray $\mathbf{r}(t) = \mathbf{o} + t\mathbf{d}$, with a camera center $\mathbf{o}$, viewing direction $\mathbf{d}$, is integrated from near bound $t_{n}$ and far bound $t_{f}$:
\begin{linenomath}
\begin{align}
\mathcal{C}(\mathbf{r};\theta,\mathcal{W},&\hat{\boldsymbol{\alpha}}) = \int_{t_n}^{t_f}\boldsymbol{\sigma}_{\theta}(\mathbf{r}(t))\cdot \mathbf{c}_{\theta}(\mathbf{r}(t),\mathbf{d})\cdot T(t)dt \\
& T(t)=exp\bigg{(}\int_{t_n}^{t} \boldsymbol{\sigma}_{\theta}(\mathbf{r}(s))ds\bigg{)}.
\label{eq:vr}
\end{align}
\end{linenomath}
Along the rays casted
through each pixel, emotional talking head is computed by  volume rendering process which accumulates the sampled density $\boldsymbol{\sigma}_{\theta}$ and RGB values $\mathbf{c}_{\theta}$, predicted by the implicit function $\theta$.  $\mathcal{W} = \{R, \omega \}$ is the predicted rigid pose parameters of the face, containing a rotation matrix $R \in \mathbb{R}^{3\times 3}$ and a translation vector $\omega \in \mathbb{R}^{3\times 1}$. 

To obtain final synthesized output, we leverage  face parsing  \cite{bigsec} for separating torso part of talking person and background image. Note that  we  focus on facial emotion so that we does not discuss the performance of torso part synthesis. The final outcome frame is the concatenation of rendered head part, segmented torso part and arbitrary background.

\begin{figure*}[t]
\begin{center}
\includegraphics[width=0.9\linewidth]{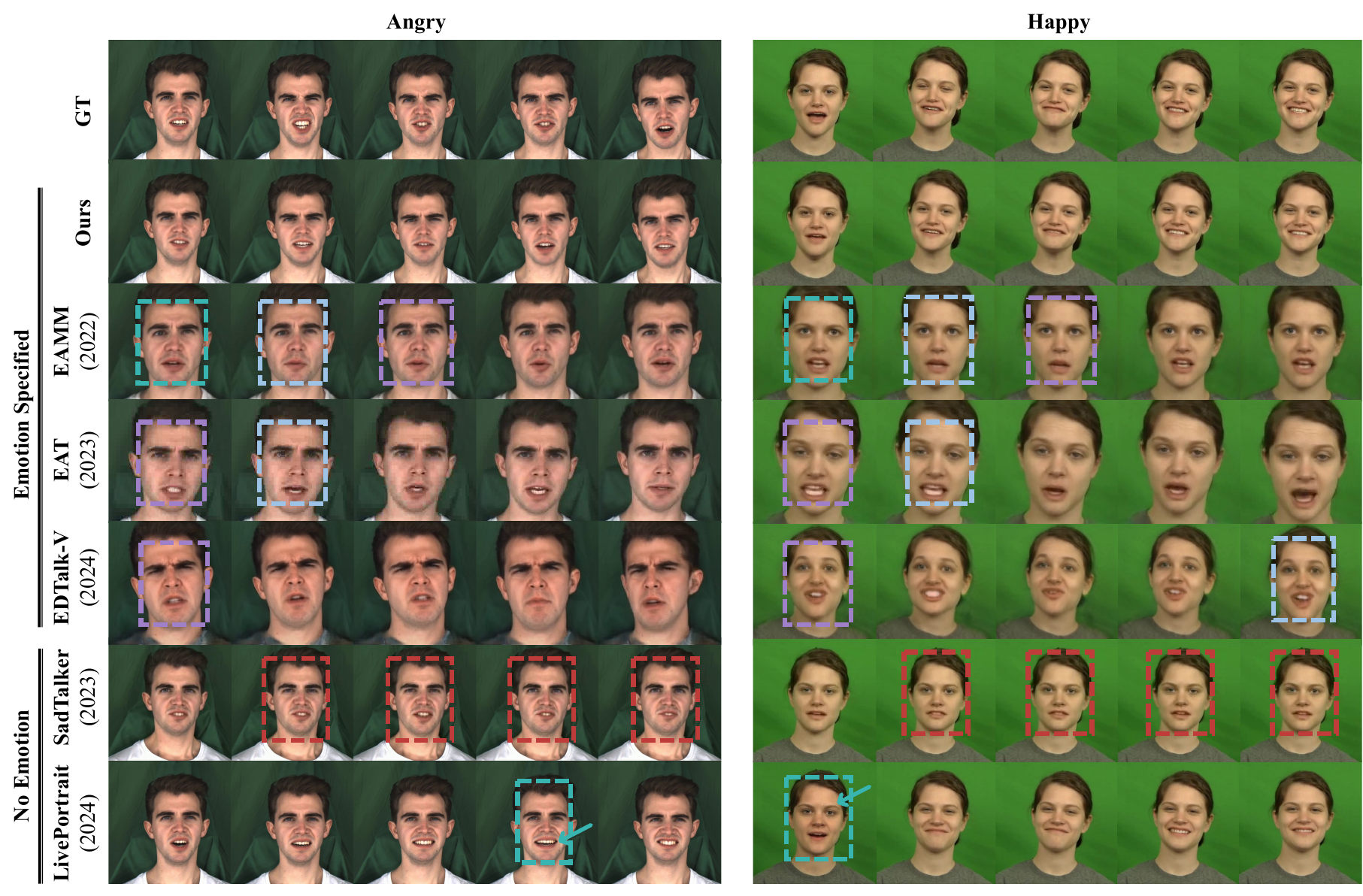}
\end{center}
\caption{Visualization comparisons of generated frames with state-of-the-art methods. See the illustration of color in Sec.\ref{quaa}.   }
\label{fig:commmp1}
\vspace{-0.15in}
\end{figure*}

\subsection{Training}

\noindent \textbf{Many-to-one Audio-expression Module}. We aim to train a universal audio-expression module that can estimate  expression parameters $ \boldsymbol{\alpha}$ from arbitrary  voice input. To this end, we leverage contrastive learning that audio features from different individuals are assigned to emotional parameters for same $z$:
\begin{linenomath}
\begin{align}
\mathcal{L}^{const}_{reg}=\|\boldsymbol{\alpha}_{\boldsymbol{\upsilon}}^{A} & -\hat{\boldsymbol{\alpha}}_{\boldsymbol{\upsilon}}^{A}\|_2  + \rho \|\boldsymbol{\alpha}_{\boldsymbol{\upsilon}}^{\bar{A}}-\hat{\boldsymbol{\alpha}}_{\boldsymbol{\upsilon}}^{A}\|_2  \\
& \hat{\boldsymbol{\alpha}} = \tilde{\boldsymbol{\alpha}} + \tau \mathbf{w}_{z}^{T},
\end{align}
\end{linenomath}
where  $\hat{\boldsymbol{\alpha}}_{\boldsymbol{\upsilon}}^{A}$ refers to the post-refined  expression parameters  given audio feature from individual $A$, $\rho$ is the hyperparameter that control the degree for learning universal facial expression parameters, empirically set as $0.5$.  $\textbf{w}^{T}_{z}$ is the trained hyperplane that separate expression parameters of emotion tag $z$ from the rest emotion tags for expression refinement. During training stage,   $\boldsymbol{\alpha}_{\boldsymbol{\upsilon}}^{\bar{A}}$ is randomly sampled from training set that belongs to another individual. Thus the audio-expression module is able to estimate emotion parameters from  different individual for same emotion tag $z$.

\noindent \textbf{Reconstruction Loss}.  
We provide 2-stage reconstruction loss $\mathcal{L}$ to train implicit function coarsely and finely:
\begin{align}
\mathcal{L}= \mathcal{L}_{photo} + \mathcal{L}_{refine}.
\end{align}
Specifically, during  first half of  training stage, let $\mathbf{I}_r = \sum_{\mathbf{r}} \mathcal{C}(\mathbf{r})$ be the rendered image  and $\mathbf{I}_g $ be the groundtruth image, we minimize the photo-metric error by $\mathcal{L}_{photo}  =\|\mathbf{I}_r-\mathbf{I}_g\|_2$ that  coarsely reconstructs target person. Next, we utilize geometry parameters $\boldsymbol{\beta}_{\mathbf{id}} \in \mathbb{R}^{50}$  \cite{guo2020towards} to ensure  implicit function  for better  representation of specific individual. Besides, since different person may show different expression for same emotion tag, we use CLIP \cite{clip_short} loss to guide the generated frame to be semantically consistent with required emotion. In detail, given the   emotion tag text $emo$, we have:
\begin{linenomath}
\begin{align}
\mathcal{L}_{refine} = \underbrace{- E_I(\mathbf{I}_r)^T E_T(emo)}_{{\mathcal{L}_{cord}}} + \underbrace{\|\boldsymbol{\beta}_{\mathbf{id}}-\hat{\boldsymbol{\beta}}_{\mathbf{id}}\|_2}_{{\mathcal{L}_{shape}}},
\end{align}
\end{linenomath}
where $ E_{I}$ and $ E_{T}$ refer to  image encoder  and text encoder  in CLIP, respectively.
 We adopt 
$\mathcal{L}_{refine}$ in the second-half of training stage, using $\mathcal{L}_{cord}$ for target emotion text-image coordination, and $\mathcal{L}_{shape}$ for enhancing reconstruction quality of target person. Finally, the weight set  $ \{\lambda_{photo}, \lambda_{cord},\lambda_{shape}\} $ is applied to achieve training stability.



\section{Experiments}
\subsection{Experiment Setup}

\begin{table*}[t]
\centering
\setlength{\tabcolsep}{2.3mm}
\caption{
Quantitative comparison and human evaluation on MEAD dataset.}\label{tab:ssim_m}{\begin{tabular}{l|c|c|c|c|c|c|c||c|c|c}
\toprule
Methods                     & SSIM $\uparrow$            & PSNR      $\uparrow$      & M/F-LMD   $\downarrow$       & FID   $\downarrow$       & AUE  $\downarrow$         & Sync. $\uparrow$           & ER $\uparrow$            & Fidelity     $\uparrow$   & Coherence     $\uparrow$ & Empathy     $\uparrow$      \\ \midrule
SadTalker \cite{zhang2022sadtalker}                                   & 0.606          & 19.042        &  2.038$/$2.335  & 69.2        & 2.86          & 5.68           & 52.11          & 3.2      & 3.8          &  3.2        \\
LivePortrait \cite{guo2024liveportrait}                               & 0.720          & 26.161          & 1.314$/$1.572  & 59.5        & 2.05          & 6.96           & 61.82          & 3.3       & 3.5          & 2.9           \\ \midrule
EAMM       \cite{ji2022eamm}                                & 0.610          & 18.867          & 2.525$/$2.814    & 31.3      & 2.35          & 1.76           & 31.08       &  2.7    & 2.9          & 3.5              \\EAT      \cite{Gan_2023_ICCV}                       & 0.652         & 20.007         & 1.750$/$1.668   & 34.6      & 1.48          & 7.98          & 64.40          & 2.7    & 3.8          & 4.0            \\
EDTalk-V      \cite{tan2025edtalk}                       & 0.769          & 22.771          & 1.102$/$1.060   & 38.5      & 1.56          & 
6.89           & 68.85          & 2.8  & 3.1         & 3.7              \\
\rowcolor{gray!30} Ours          &                  \textbf{0.891} & \textbf{28.262} & \textbf{1.031$/$0.981}  & \textbf{30.9} & \textbf{1.41} & \textbf{8.08} & \textbf{73.28} &   \textbf{4.6}  & \textbf{4.3}        & \textbf{4.2} \\ \midrule
Groundtruth                                    & -              & -         & -      & -              & -             & 7.36          & 79.65          & 4.9              & 4.7       & 4.4     \\ \bottomrule

\end{tabular}}
\vspace{-0.05in}
\end{table*}

\begin{table*}[t]
\centering
\setlength{\tabcolsep}{2.3mm}
\caption{
Quantitative comparison and human evaluation on  CREMA-D dataset.}\label{tab:ssim_d}{\begin{tabular}{l|c|c|c|c|c|c|c||c|c|c}
\toprule
Methods                     & SSIM $\uparrow$            & PSNR      $\uparrow$      & M/F-LMD   $\downarrow$       & FID   $\downarrow$       & AUE  $\downarrow$         & Sync. $\uparrow$           & ER $\uparrow$            & Fidelity     $\uparrow$   & Coherence     $\uparrow$ & Empathy     $\uparrow$      \\ \midrule
SadTalker \cite{zhang2022sadtalker}                                   & 0.688          & 25.572          &  2.851$/$2.156  & 63.7        & 2.01          & 5.99           & 69.59          & 3.4       & 3.7          & 3.8        \\
LivePortrait \cite{guo2024liveportrait}                               & 0.873          & 28.985          & 1.411$/$1.503   & 33.6        & 1.94          & 6.02           & 70.13          & 3.5       & 3.6          & 3.6           \\ \midrule
EAMM       \cite{ji2022eamm}                                & 0.650          & 21.910         & 2.751$/$2.824    & 35.6      & 2.19          & 3.22           & 40.72       &  3.2    & 3.1          & 3.3             \\
EAT      \cite{Gan_2023_ICCV}                       & 0.794          & 23.712          & 1.925$/$1.791  & 32.3      & 1.98          & 5.70           & 61.76           & 3.1    & 3.7        & 3.5           \\
EDTalk-V      \cite{tan2025edtalk}                        & 0.753          & 27.641          & 1.228$/$1.137   & 40.5      & 1.81          & 5.57           & 54.46          & 3.0  & 2.9          & 3.2              \\
\rowcolor{gray!30} Ours          &                  \textbf{0.914} & \textbf{31.621} & \textbf{1.191$/$1.035}  & \textbf{29.3} & \textbf{1.55} & \textbf{6.11} & \textbf{75.98} &   \textbf{4.5}  & \textbf{4.7}        & \textbf{4.3} \\ \midrule
Groundtruth                                    & -              & -         & -      & -              & -             & 7.09          & 82.51          & 4.9              & 4.8          & 4.8     \\ \bottomrule

\end{tabular}}
\vspace{-0.15in}
\end{table*}

\noindent \textbf{Dataset}. We apply the multi-view emotional audio-visual dataset (MEAD) \cite{kaisiyuan2020mead} and the crowd-sourced emotional multimodal actors dataset (CREMA-D) \cite{CaoCA}. MEAD contains 60 expressive actors and actresses who express eight distinct emotions at three varying intensity levels. CREMA-D \cite{CaoCA} consists of 7,442 original clips from 91 actors, each with four different emotion levels. \textbf{Baseline Methods}. 
We compare EmoHead against non-emotional methods and emotion-specific methods. For non-emotional methods, we choose SadTalker \cite{zhang2022sadtalker} and LivePortrait \cite{guo2024liveportrait}. We use the original test video as the driving source and the first frame as the input emotion condition. For emotion-specific methods, we preset one target emotion and compare the state-of-the-art methods EAMM \cite{ji2022eamm}, EDTalk \cite{tan2025edtalk}, and EAT \cite{Gan_2023_ICCV}. We implement all the default hyperparameters for the baseline methods. \textbf{Metrics}.  
For the generated video, we use PSNR and SSIM to quantify reconstruction quality. We also evaluate the generated frames using the FID \cite{heusel2017gans}, which measures image quality and diversity. Additionally, we employ the action unit error (AUE) \cite{8373812} and face landmark distances (F-LMD) for evaluating the accuracy of face motion; the SyncNet score (Sync.) \cite{Chung16a} and mouth landmark distances (M-LMD) \cite{chen2019hierarchical} for assessing lip synchronization; and the accuracy of emotion recognition (ER) \cite{meng2019frame} in frames to evaluate the emotion consistency of generated video with target emotion.


\subsection{Quantitative Analysis}

\noindent \textbf{Comparison with Baseline Methods}. Tab.~\ref{tab:ssim_m} and \ref{tab:ssim_d} shows the reconstruction quality produced by our proposed framework. Our method achieves the highest  SSIM and PSNR scores, resulting in high-quality talking head videos. The AU error and F-LMD from proposed method consistently show the highest results on both datasets, illustrating that our approach also generates more realistic face motion and accurate  expression than emotion-aware method EAMM, EDTalk-V and EAT.  It indicates that the proposed audio-expression module successfully enhances the correlation between facial movement and audio input. As for the accuracy comparison of facial expression generation, we calculate mean accuracy of across all the emotions in the test set. We surpass the state-of-the-art method EDTalk  which generates accurate target emotion of talking head. This is due to the emotion-specific hyperplane disentangles expression parameters at low-dimension space and regularize NeRF to  increase the semantic representation of emotional talking head synthesis.


\noindent \textbf{Ablation Studies}.
Please kindly refer to Appendix (D\&F) for the audio expression alignment, expression refinement and loss implementation.

\noindent \textbf{Human Evaluation}.  We conduct generation quality and emotion-based pairwise preference test for EmoHead. For a given pair of generated video and emotion tag, we assign crowd sourcing workers to annotate the video with a score on scale of 0 to 5 and report average scores. We define three aspects for evaluation: Fidelity, Coherence, and Empathy. Fidelity refers to the reconstruction quality of generated video. Coherence assesses the consistency of emotion with expression. Empathy refers to the degree to which each video encourages the user to respond with emotion.  Tab.~\ref{tab:ssim_m} and \ref{tab:ssim_d} show that the generated talking heads from our model exhibit higher fidelity and coherence, indicating that refined expression parameters lead to more accurate emotion representation. Our model also outperforms all the baselines in terms of empathy, evoking user for a more emotional experience.



\begin{figure}[t]
\begin{center}
\includegraphics[width=0.9\linewidth]
{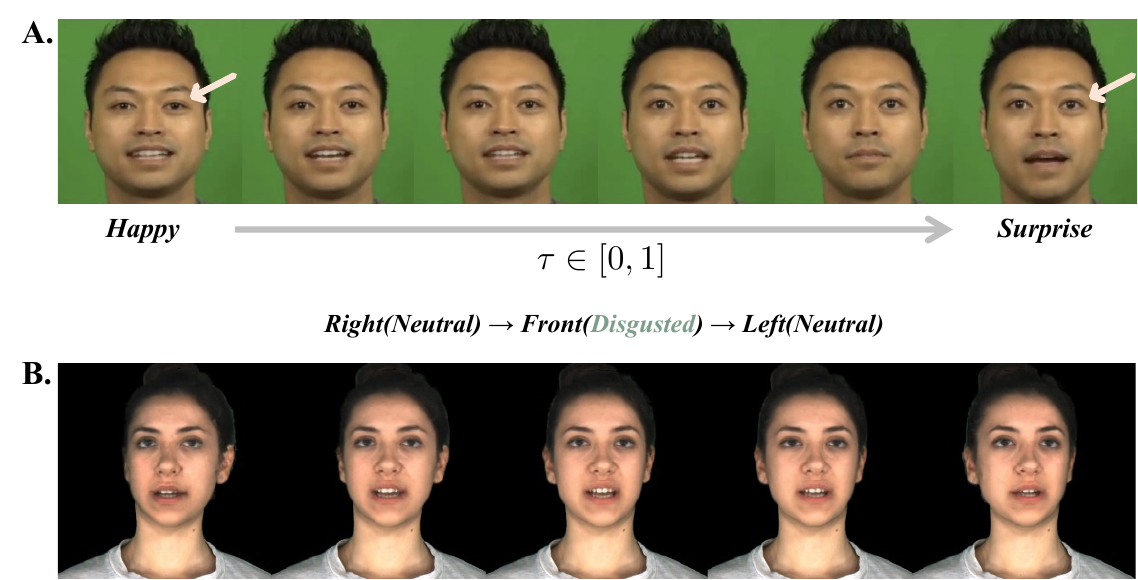}
\end{center}
\caption{(A) Continuous expression manipulation in talking stage. (B) Reconstruction result of unseen views and varying emotion. Please refer to the supplementary video.}
\label{fig:cont}
\vspace{-0.15in}
\end{figure}

\subsection{Qualitative Analysis} \label{quaa}
\noindent \textbf{Comparison with Baseline Methods}.   
We provide visual examples of specific emotions to demonstrate the efficacy of the proposed method. Emotional talking heads are generated using input audio clips corresponding to each respective emotion for all methods. In Fig.~\ref{fig:commmp1}, our method shows the best fidelity for talking head synthesis compared to the other methods. We use different colors to highlight the shortcomings of the baselines: (1) Generated output from SadTalker is limited to the source image, resulting in an output sequence that \textcolor{purple}{lacks of expression  variability}. (2) LivePortrait \textcolor{teal}{generates inappropriate emotions} since the mouth region displays unexpected shapes and angles that do not semantically correspond to anger. (3) EAMM, EAT, and EDTalk-V \textcolor{violet}{are unable to generate the same identity as the target person}. (4)  EAMM, EAT, and EDTalk-V \textcolor{babyblue}{fail to generate high-fidelity talking faces}. Our method successfully generates consistent facial emotion without showing expression collapse.

\noindent \textbf{Refinement Efficacy}. 
Please kindly refer to Appendix (B\&E) for the efficacy of the proposed methods in Sec.\ref{sec:4} \& \ref{sec:emosec3}.




\noindent \textbf{Continuous Expression} 
As shown in Fig.~\ref{fig:cont}.A, our method can produce continuous facial expression changes by Eq.~\eqref{eq:intecthp} that linearly interpolates between two expression parameters from different emotion tags during the talking stage, which cannot be accomplished by the baselines. Additionally, based on the neural radiance field, our method can generate talking head videos from unseen views with expression variation, which facilitates continuous talking head syntheses of different views and emotions, as depicted in Fig.~\ref{fig:cont}.B. Please refer to the supplementary video.

\section{Conclusion}
In this paper, we present emotional talking head generation using emotion tag and refined expression parameter from a morphable face model. We provide a novel framework that includes an audio-express module for predicting expression parameters from audio input and an expression rendering module for generating person-specific talking head videos. The predicted facial expression parameters can be manipulated using pre-trained hyperplanes for expression refinement. Experimental results show that the proposed framework is effective for emotion-controllable talking head synthesis.

\noindent \textbf{Acknowledgments.} This work is supported in part by NSFC Project (62176061) and Shanghai Municipal Science and Technology Major Project
(No.2021SHZDZX0103).

\bibliographystyle{IEEEbib}
\bibliography{icme2025references}


\clearpage

\section*{Appendix of EmoHead}

\begin{table*}[t]
\centering
\setlength{\tabcolsep}{1.4mm}
\caption{Comparison with Baselines in terms of emotion manipulation. ``-'' denotes that the method is not able to achieve the function. ``3DMM Exp.''  refers to the expression coefficients in 3DMM \cite{guo2020towards} \label{tab:compother}}{\begin{tabular}{@{}l|cccc@{}}
\toprule
\textbf{Method} & \textbf{Manipulation Source} & \textbf{Varying Within Two Emotions} & \textbf{Varying Views} & \textbf{Varying Different Emotions In Talking} \\ \midrule
LivePortrait, 2024  \cite{guo2024liveportrait}  & -                            & -                            & -                     & -                                  \\
SadTalker, 2023   \cite{zhang2022sadtalker}    & -                            & -                            & -                     & -                                  \\
EVP, 2020      \cite{ji2021audio}       & Emotion Encoding (audio)                         & \ding{52}                             & -                     & -                                  \\
EAMM, 2022     \cite{ji2022eamm}       & -                            & -                            & -                     & -                                  \\
EAT, 2023     \cite{Gan_2023_ICCV}        & CLIP (text)                          & -                            & -                     & -                                  \\
EDTalk, 2024   \cite{tan2025edtalk}       & Emotion Encoding (audio \& text)                        & \ding{52}                          & \ding{52}                  & -                                 \\ \midrule
Ours            & 3DMM Exp. (hyperplane)                   & \ding{52}                          & \ding{52}                   & \ding{52}                                \\ \bottomrule
\end{tabular}}
\end{table*}

\subsection{Entangled Expression Parameters in 3DMM}

The 3D Face Morphable Model (3DMM) has been a fundamental area of natural face reconstruction, first proposed by Blanz et al. \cite{blanz1999morphable} in 1999. Initially developed as a linear model using the PCA algorithm,  3DMM is capable of representing the shape and texture of a 3D face model. Subsequent research has focused on enhancing its performance through the utilization of larger 3D face datasets \cite{Pascal2009,Li2020Formation,Linchao2020High}.

\begin{figure*}[t]
\begin{center}
\includegraphics[width=0.93\linewidth]{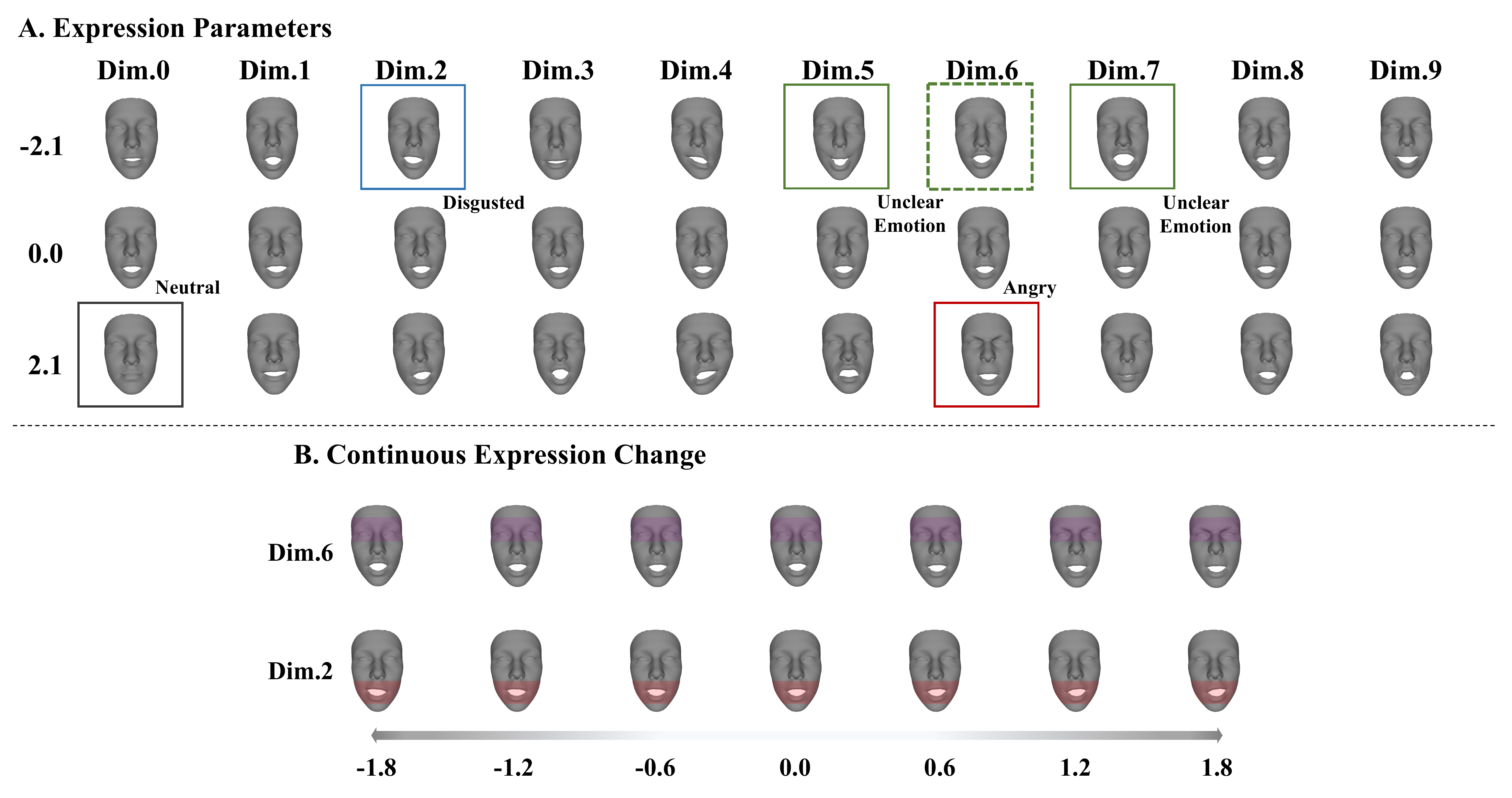}
\end{center}
\caption{Figure A displays the monocular reconstruction by changing the value of each dimension for the expression parameters of 3DMM. Emotions can hardly be reconstructed using facial expression parameters. Most dimensions represent ``unclear'' emotions, as depicted in the green boxes. Figure B shows the continuous expression change in a single dimension through the range $ [-1.8, 1.8]$.   }
\label{fig:long}
\end{figure*}

Recent 3D face datasets exhibit improved diversity in expression parameters.  \cite{chen2024morphable}   demonstrates superior generalization in facial expression fitting and other datasets \cite{yang2020facescape,wang2022faceverse} with rich facial expressions has been collected to enhance the incorporation of facial expression bases into 3DMM.  The extracted facial expression parameters from a 2D avatar satisfies continuous facial expression change through a given range. Recently, due to the continuous expression change, depicted in Figure~\ref{fig:long}.B, previous work \cite{yao2022dfa,Gan_2023_ICCV,ye2024real3d} leverage the estimated expression parameters to represent emotion in talking head video.


However, facial expression parameters do not semantically represent target emotion. Take the reconstruction result from 3DDFA \cite{guo2020towards} as example, In Figure~\ref{fig:long}.A, by changing the value of a single dimension, we find that only a few dimensions can  display interpretable facial expressions with respect to target emotion. For instance, the ``angry''  emotion is shown only by the eye region after setting $\text{Dim.6} = 2.1$, as indicated by the \textcolor{red}{\textbf{red}} box in Figure~\ref{fig:long}.A. Nevertheless, most dimensions do not represent specific emotions, as shown by the \textcolor{teal}{\textbf{green}} boxes. For example, after  changing ``Dimension 6'' from $\text{Dim.6} = 2.1$ to $\text{Dim.6} = - 2.1$, the ``angry'' emotion is  emotion is lessened but shifted into an unclear emotion. Those unclear emotions lead to ``\textbf{expression collapse}'' phenomenon and potentially decrease generation quality of emotional talking head.   In this work, we aim to disentangle  the facial expression parameters to ensure specific emotion of talking head, which is crucial for generating high-quality and emotionally natural  talking head video.  



\subsection{Emotion Hyperplane}
\noindent \textbf{Pre-training  hyperplane}. In this section, we introduce the training details for the hyperplane designed to distinguish facial expressions corresponding to desired emotions from other emotions. Firstly, we employ 3DDFA \cite{guo2020towards} to transform video frames of the target character in the training dataset into facial expression parameters. After the transformation into facial parameters, we label all the parameters based on their emotional tags. Subsequently, using these emotional labels, we train a hyperplane using SVM in Scikit-learn \cite{pedregosa2011scikit}. To ensure that the expression parameters effectively adapt to varying mouth openings during speech, we utilize facial landmarks to categorize five arithmetic groups, defined by the maximum and minimum of the MAR (mouth aspect ratio). During training, we evenly sample and balance the quantities across these five groups. After obtaining the trained hyperplanes, we apply them to adjust the expression parameters to the target emotion, as depicted in the lower-left part of Figure 2 (in the main paper).

\begin{figure*}[t]
\centering
\includegraphics[width=0.95\linewidth]
{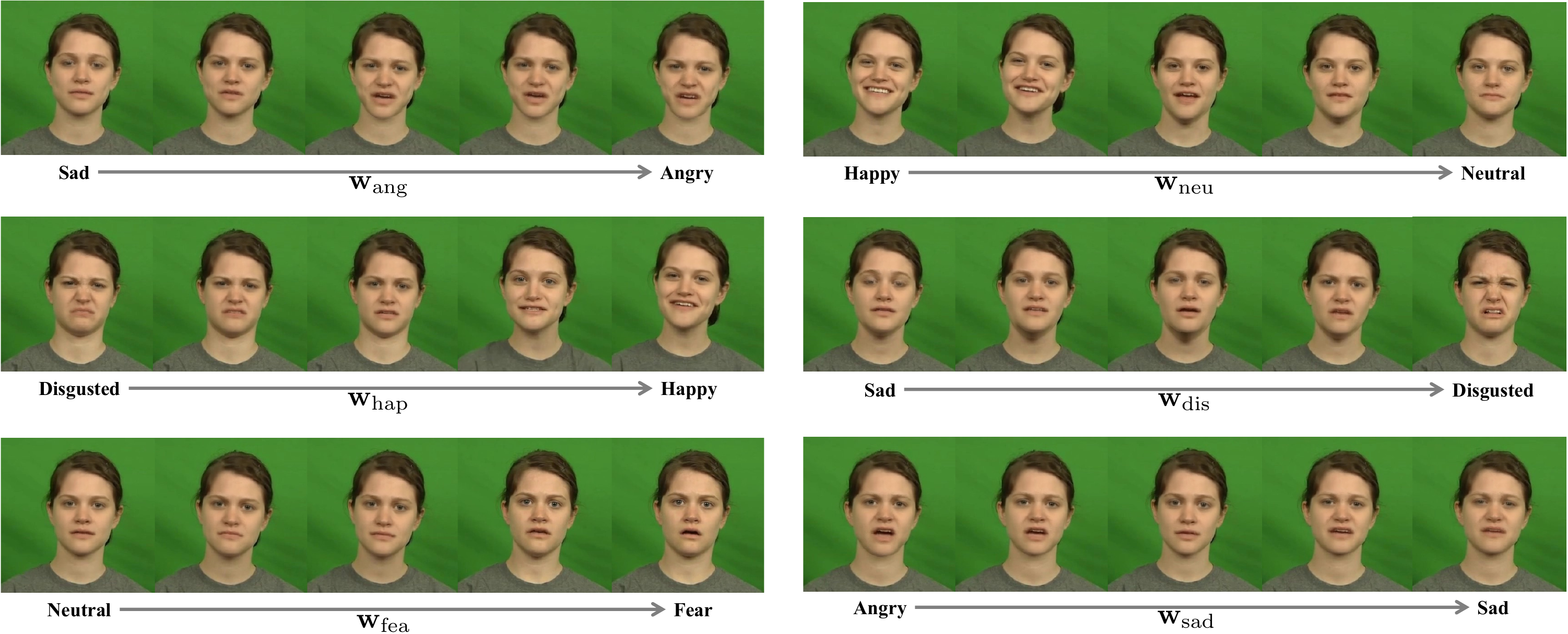}
\caption{Talking-stage visualization of  probing expression parameters through trained hyperplane for all six emotions in CREMA-D dataset.}
\label{fig:view}
\end{figure*}

\begin{figure*}[t]
\centering
\includegraphics[width=0.7\linewidth]
{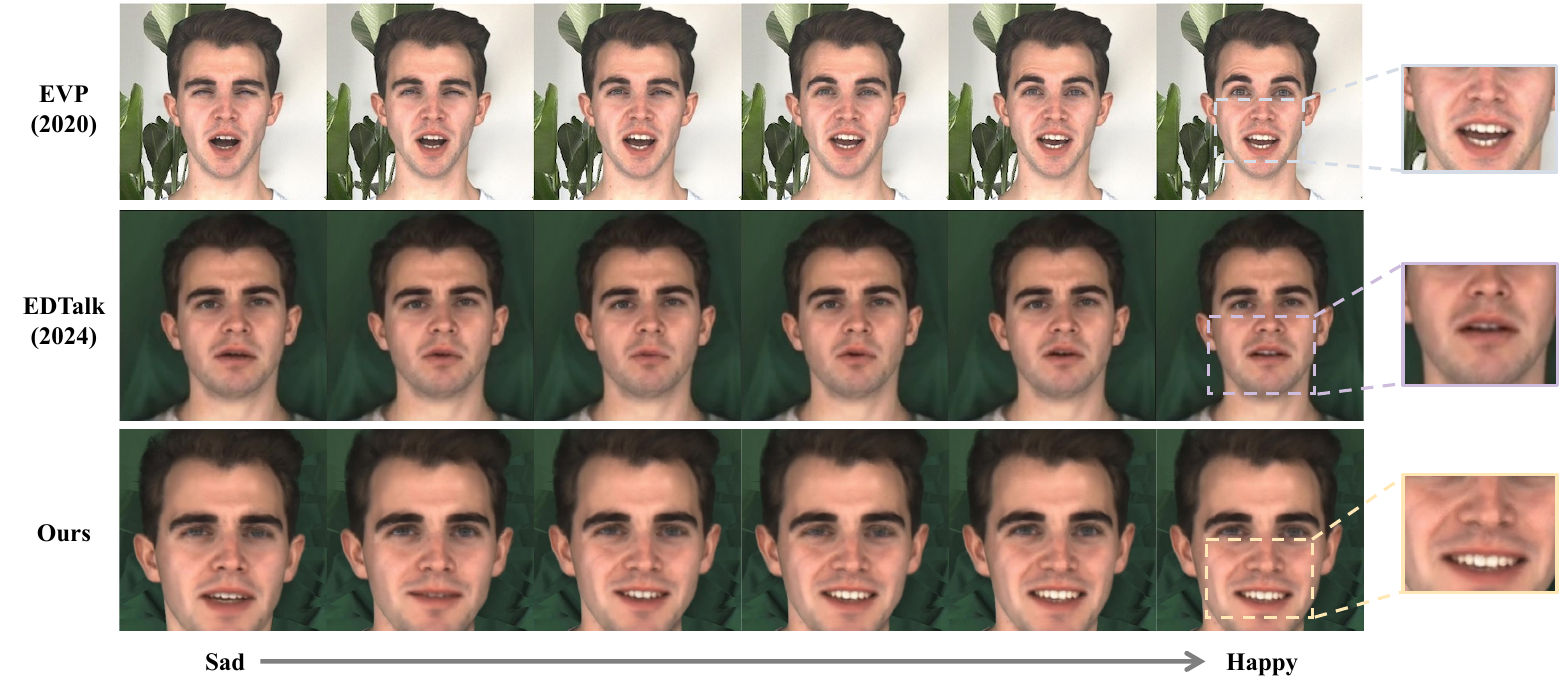}
\caption{Comparison of expression editing between two emotions. We enlarge the mouth region for better visualization. }
\label{fig:cont_view}
\end{figure*}

\noindent \textbf{The efficacy of the emotion hyperplane}.  In Table~\ref{tab:compother}, we compare our method with baseline emotional talking head synthesis in terms of emotion manipulation. To the best of our knowledge, we are the first to combine emotion variations in talking, which benefits emotional human-machine interaction. 
Here we conclude the efficacy and novelty of our proposed method:

(1) Through this emotion refinement method, we can successfully manipulate expressions based on their normal vectors for talking heads. For instance, if we aim to make an expression appear happier, we can move along the normal vector of happiness and apply a weighted coefficient in the opposite direction. We provide all six emotion variations in the CREMA-D dataset, as depicted in Figure~\ref{fig:view}. Additionally, Figure~\ref{fig:cont_view} demonstrates that the edited emotions are more semantically accurate than those of the baselines (EVP \cite{ji2021audio} and EDTalk \cite{tan2025edtalk}), particularly in the mouth regions. Consequently, through this plug-and-play method, we  enhance the representation of the facial parameters obtained after prediction from audio input to expression parameters. Please refer to the emotion variations within two emotions  and views in the supplementary video (approximately at the 3:45 mark).

\begin{figure*}[t]
\begin{center}
\includegraphics[width=0.8\linewidth]
{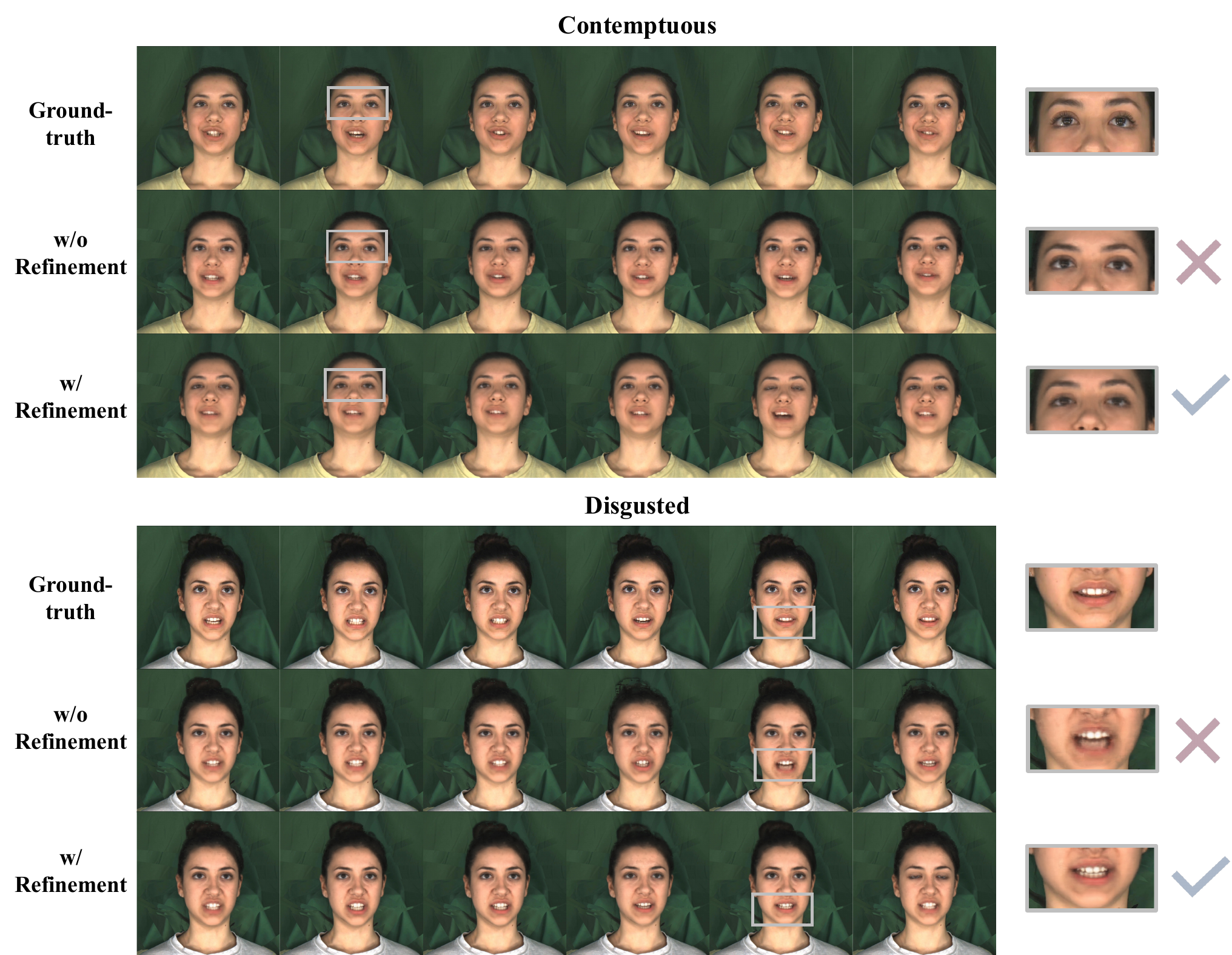}
\end{center}
\caption{Additional results of expression refinement. Gray boxes display  the refined region of expression. }
\label{fig:add2}
\end{figure*}

(2) The emotion hyperplane improves the reconstruction quality of representing specific emotions. Figure~\ref{fig:add2} illustrates two refined emotions of the same character. Our proposed expression refinement method, along with the emotion hyperplane, semantically enhances generation quality and alleviate ``expression collapse'' phenomenon, as shown in the gray boxes of Figure~\ref{fig:add2}. Surprisingly, the character's eyes appear smaller in our method for the emotion ``Contempt", which performs even better than the ground truth.

(3) The proposed hyperplane is beneficial for generating talking heads of two different emotions in talking stage, which can not be accomplished by previous work. Specifically, we adopt the trained hyperplanes from two different emotions, denoted as $\textbf{w}_1$ and $\textbf{w}_2$. During the inference stage, we substitute $\textbf{w}_1$ with $\textbf{w}_2$ in the pre-determined time and linearly probe them with the weight  $\tau$. We demonstrate that the generated face can successfully represent the combined emotion with the semantic expression in a single video clip. Please refer to the emotion variations during the talking stage in the supplementary video, which is located in the application section (approximately at the 4:45 mark).

\subsection{Implementation and Training Details}
\begin{figure*}[t]
\begin{center}
\includegraphics[width=0.81\linewidth]{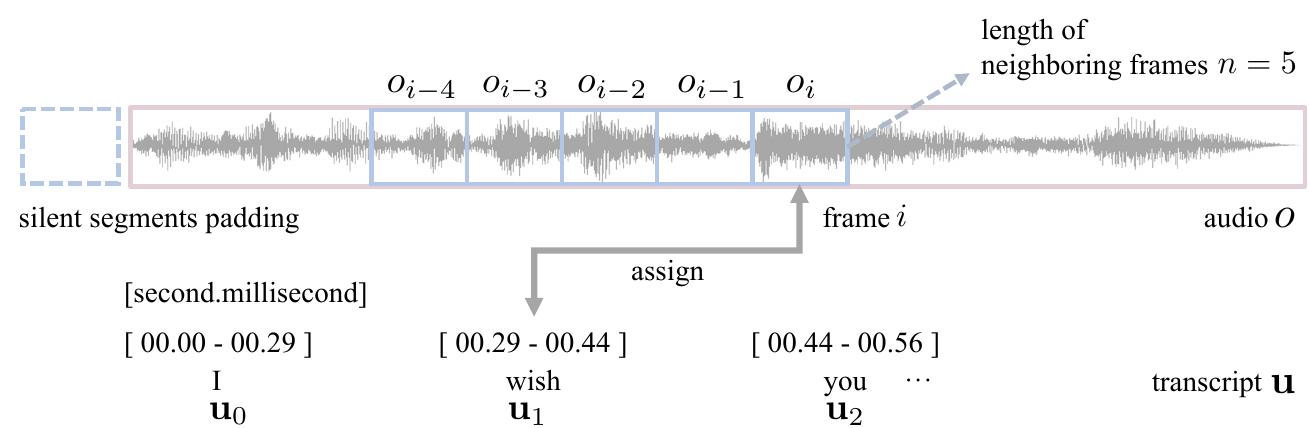}
\end{center}
\caption{Implementation details for  aligned audio and transcript inputs. We assign the word $\textbf{u}_j$ to  audio $o_i$ based on timestamp.  }
\label{fig:neighbour}
\end{figure*}

\noindent \textbf{Implementation Details}. First, 
in the Audio-Expression module, we are inspired by the Fused Attention module of \cite{Shuhan_WWW24} but Emohead has different domains and series window of model input. The input data consists of audio $o$ and the frames per second (FPS) is set to 25 to synchronize the audio input with the video frames. Concurrently, we employ FunASR \cite{gao2023funasr} to transcribe audio to text. This transcription is timestamped, allowing us to correlate the audio with its corresponding timestamps. For instance, at the $i$-th frame, we obtain an audio segment  $o_i$ and the corresponding text segment $\textbf{u}_j$. Furthermore, within the fusion function of $\mathcal{F}$, the length $n$ of neighboring frame  frames is set to $5$. Since the initial audio segments have an insufficient number of neighboring frames, we employ silence segment padding to fill the gaps, as illustrated in Figure~\ref{fig:neighbour}.

Then, we provide Table~\ref{tab:Desc} to display the detailed feature dimensions in our framework. Note that the dimension of the hyperplane $\textbf{w}$ corresponds to the expression coefficients of 3DMM \cite{guo2020towards}, where $\boldsymbol{\alpha} \in \mathbb{R}^{10} $. For the two feedforward networks in the audio expression module (Sec.III.A of the main paper), we use the same network architecture $(512,256,256,128,k)$, where $k=10$  for predicting expression parameters $\boldsymbol{\alpha}$ and $k=1$  for predicting hyperplane weights $\boldsymbol{\tau}$. 

For the rendering module, we choose the instant-ngp version of AD-NeRF \cite{tang2022real} to achieve better efficiency of training and inference. The implicit function $\theta$ has the same network architecture in \cite{mildenhall2020nerf}.

\noindent \textbf{Training Details of Two datasets}. For both datasets, 
we use PyTorch and NVIDIA A100 to conduct all the experiments. To train the audio-expression module, all the parameters are updated with Adam optimizer  with initial learning rate $0.0005$. We train audio-expression module for 20k iterations and rendering module for 200k iterations. We apply $\mathcal{L}_{refine}$ after 100k iterations.  To train the implicit function in rendering module, we set 200,000 iterations for both datasets. In the first 100,000 iterations (first-half stage), we implement reconstruction loss with the Adam optimizer and a linear learning schedule. Then, we employ $ \mathcal{L}_{cord} $ and $ \mathcal{L}_{shape} $ at the second-half training stage. Within  $ \mathcal{L}_{cord}$, the CLIP image encoder $E_{I}$ is CLIP-RN50 and text encoder $E_{T}$  is the network fine-tuned on GPT-2 \cite{clip_RadfordKHRGASAM21}. The weights of loss are set to $ \lambda_{photo} =1$, $ \lambda_{cord} = {1e-3}$  and $ \lambda_{shape} = {1e-9} $, correspondingly, to ensure a steady and effective training process. 

Differently, for the MEAD dataset, we crop the training videos to dimensions of $512\times 512$, with each individual providing 10 minutes of training data and 2 minutes of testing data. For the CREMA-D dataset, which is smaller than MEAD, the training videos are cropped to $340 \times 340$, with 2 minutes of video for training data and 1 minute of video for testing data.



\begin{table*}[t]
\centering
\setlength{\tabcolsep}{4.1mm}
\caption{
Quantitative comparison of ablation studies on MEAD dataset.}\label{tab:ssim-abla}
{\begin{tabular}{l|c|c|c|c|c|c|c}
\toprule
Settings                     & SSIM $\uparrow$            & PSNR      $\uparrow$      & M/F-LMD   $\downarrow$       & FID   $\downarrow$       & AUE  $\downarrow$         & Sync. $\uparrow$           & ER $\uparrow$             \\ \midrule
w/o audio expression alignment                                 & 0.730          & 25.191          & 1.149$/$1.082  & 89.6        & 1.95          & 6.19           & 63.44  
\\ 
w/o hyperplane refinement                             & 0.872          & 27.410          & 1.291$/$1.203   & 59.2        & 1.86          & 7.25           & 61.29                          \\ Ours    &                  \textbf{0.891} & \textbf{28.262} & \textbf{1.031$/$0.981}  & \textbf{30.9} & \textbf{1.41} & \textbf{8.08} & \textbf{73.28}   \\ \midrule
     Ours (FaceVerse)    \cite{wang2022faceverse}                             & 0.872          & 27.740          & 1.175$/$1.225   & 43.3      & 1.52          & 7.39           & 61.88                              \\
Ours (MorphableDiffusion)    \cite{chen2024morphable}                          & 0.886          & 27.731          & 1.130$/$1.066   & 38.1      & 1.55          & 7.01           & 70.31                           \\
Ours (3DDFA)    \cite{guo2020towards}      &                  \textbf{0.891} & \textbf{28.262} & \textbf{1.031$/$0.981}  & \textbf{30.9} & \textbf{1.41} & \textbf{8.08} & \textbf{73.28}   \\ \midrule
     $d,d_h = 256,256$                             & 0.717          & 27.018          & 1.092$/$1.119    & 33.7      & 1.66          & 6.58           & 66.26                              \\ $d,d_h = 512,512$                            & \textbf{0.891}          & \textbf{28.262}         & \textbf{1.031$/$0.981}    & \textbf{30.9}     & 1.41          & 8.08      & \textbf{73.28}                               \\ 
$d,d_h = 768,768$                          & 0.872          & 28.155          & 1.087$/$1.121   & 31.3      & \textbf{1.36}           & 7.05           & 72.57                          \\
$d,d_h = 1024,1024$      &                  0.852 & 27.719 & 1.093$/$1.139  & 33.5 & 1.44  & \textbf{7.96} & 70.61   \\ \midrule $n = 3 $      &                  0.803 & 26.114 & 1.072$/$1.087  & 34.6 & 1.56  & 6.77 & 69.31   \\ $n = 5$      & \textbf{0.891}          & \textbf{28.262}         & \textbf{1.031$/$0.981}    & \textbf{30.9}     & \textbf{1.41}          & \textbf{8.08}      & 73.28     \\ $n = 8 $      &                  0.869 & 27.256 & 1.064$/$0.991  & 31.6 & 1.49  & 7.58 & 72.17   \\ $n = 12$      &                  0.885 & 28.223 & 1.105$/$1.076  & 32.9 & 1.45  & 7.96 & \textbf{74.42}   \\ \bottomrule

\end{tabular}}
\end{table*}

\subsection{Ablation Studies of Quantitative Analysis} 
We quantitatively evaluate the effectiveness of audio expression alignment and expression refinement:
\begin{itemize}
    \item w/o audio expression alignment:  We use simple feature fusion method, that is, we sent the concatenated feature $ [\boldsymbol{\upsilon};\textbf{e};\boldsymbol{\gamma}] $ to a typical self-attention transformer \cite{vaswani2017attention} for predicting expression parameter and  hyperplane  weights.
    \item w/o  hyperplane refinement: We does not implement expression refinement $\hat{\boldsymbol{\alpha}} = \tilde{\boldsymbol{\alpha}} + \tau \mathbf{w}_z^{T} $ for rendering talking head. Instead, the input of rendering module is $\hat{\boldsymbol{\alpha}} = [z;\tilde{\boldsymbol{\alpha}}]$, where $z \in \mathbb{N}^+$ is a integer that represent different emotion tags.
\end{itemize}
As shown in the top part of Table \ref{tab:ssim-abla}, the reconstruction quality without audio expression alignment decreases significantly, showing much lower SSIM and PSNR scores. Furthermore, it indicates that by enhancing the correlation between audio features and emotional features, we can better align audio features with the transcript, leading to improved lip synchronization and showing higher Sync and M-LMD score. Regarding expression refinement, we observe that without this refinement, the performance of emotion representation decreases, as indicated by lower AUE  and ER score. This suggests that expression refinement enhances the emotion representation for talking head synthesis.

\begin{table*}[t]
\centering
\setlength{\tabcolsep}{5.3mm}
\caption{
Details of Feature Dimensions.\label{tab:Desc}}{\begin{tabular}{@{}lll@{}}
\toprule
\textbf{Description}                              & \textbf{Encoder} & \textbf{Size} \\ \midrule
Dimension of Audio feature  $\tilde{\boldsymbol{\upsilon}} $          & HuBERT-base      & $768$           \\
Dimension of Audio Emotion   Feature  $\tilde{\boldsymbol{e}}$            & Emotion2vec-base & $768$           \\
Dimension of Text feature    $\tilde{\boldsymbol{\gamma}}$                     & LLaMA2-7B        & $4096$          \\
The Hidden Dimension for Audio   Expression Alignment and projection matrix $d,d_h$ & -                & $512,512$          \\ 
Dimension of Emotion   Hyperplane   $m$              & -                & $10$            \\ \bottomrule
\end{tabular}}
\end{table*}

In order to explore how the coefficients of the 3D Face Morphable Model affect our method, we compare the expression parameters of three prevalent models. We choose MorphableDiffusion \cite{chen2024morphable}, FaceVerse \cite{wang2022faceverse}, and 3DDFA \cite{guo2020towards} to regress the expression parameters for the audio-expression module. We implement dimensions of 100, 50, and 10 for the expression parameters of MorphableDiffusion, FaceVerse, and 3DDFA, respectively. Table \ref{tab:ssim-abla} demonstrates the similar performance of the three sets of expression parameters. This indicates that we can leverage 3DDFA as the source expression parameter model due to its higher runtime efficiency without compromising the quality of reconstruction.

In the third part of Table~\ref{tab:ssim-abla}, we compare the size of the hidden dimension for audio expression alignment, denoted as $d$. It can be observed that using dimension of $512$ significantly  achieves better performance than $256$, but there are  no significant improvements when the dimension exceeds $512$. Thus, for computation efficiency, we directly set $d=512$.


In the bottom part of Table~\ref{tab:ssim-abla}, we provide a comparison of the number of neighboring frames  
$n$ for different fused functions in audio expression alignment. The $n = 5$ configuration outperforms the others, except in the case of ER when $n = 12$. However, the lip synchronization for $n = 12$ is significantly lower than for $n = 5$, so we use $n = 5$ in our experiments.

\subsection{Efficacy of Audio Expression Alignment} 
Based on the proposed audio expression module, our method can generate emotional talking head videos with different emotion tags, regardless of the emotion conveyed in the audio input. Figure~\ref{fig:add_dv} demonstrates that we generate three distinct emotions in different individuals using an audio input with  ``fear'' emotion.  Our proposed audio expression alignment successfully maps various emotional audio inputs to the desired emotional expressions, regardless of audio emotion and gender identification. Please refer to the supplementary video for detailed generation comparison (approximately at the 2:21 mark).

\subsection{Loss Implementation} 
\begin{figure*}[t]
\begin{center}
\includegraphics[width=0.98\linewidth]
{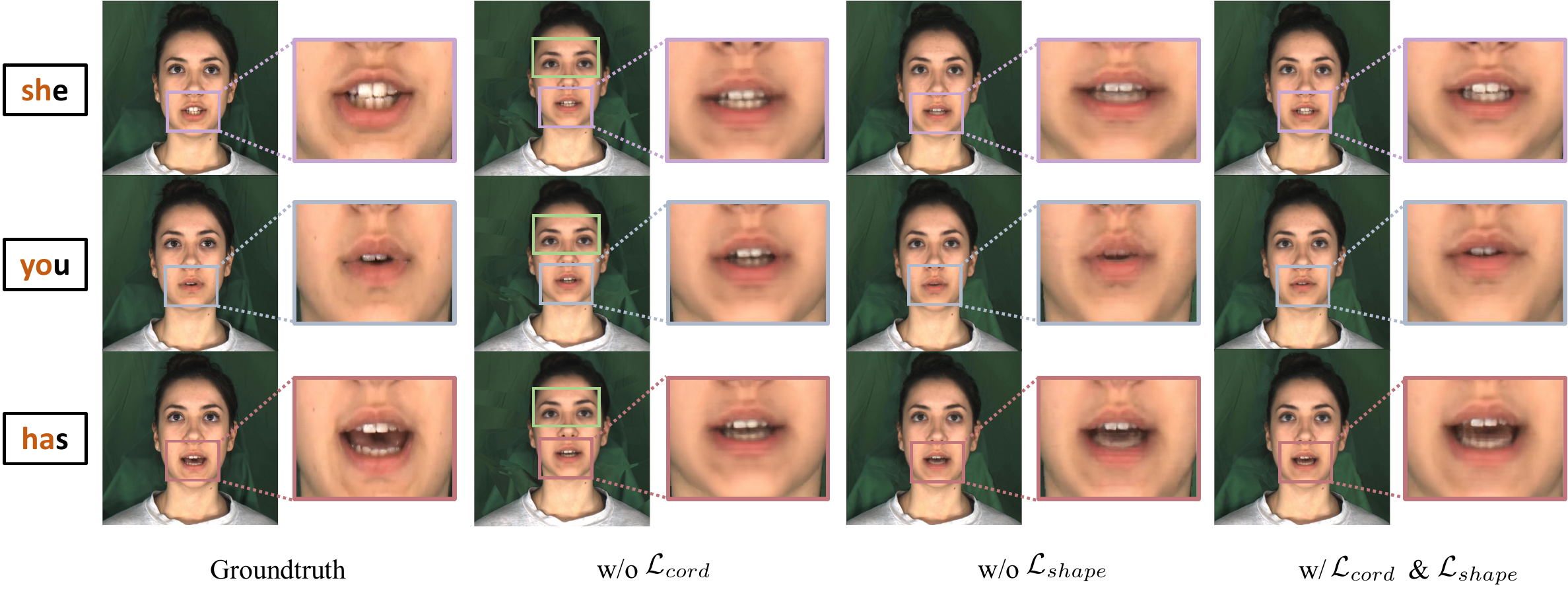}
\end{center}
\caption{Visualization of loss implementation. We present three pronunciations to compare the lips motion synthesis quality of emotional talking head. }
\label{fig:abla}
\end{figure*}

\begin{figure*}[t]
\begin{center}
\includegraphics[width=0.81\linewidth]
{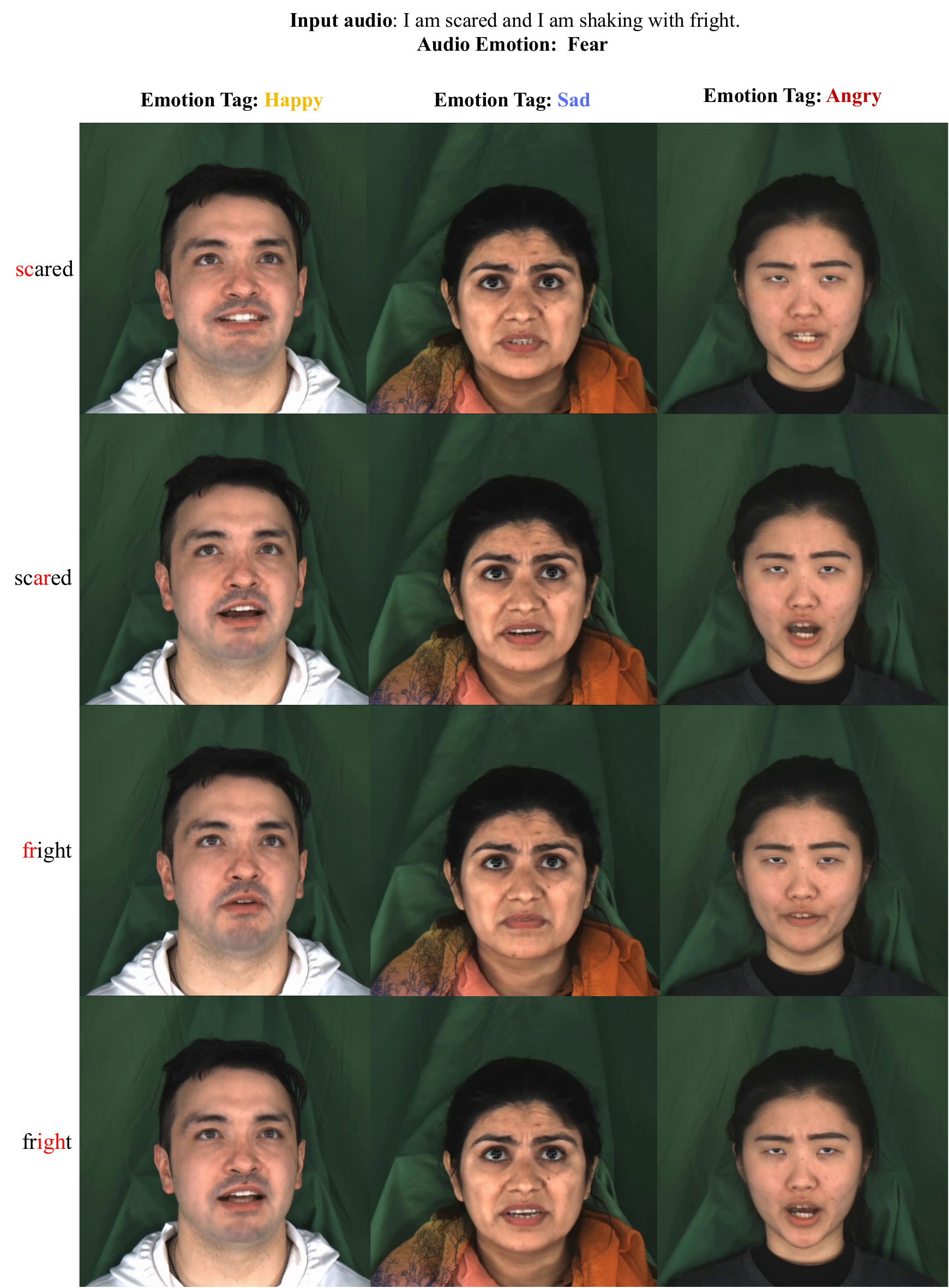}
\end{center}
\caption{Generated results of  emotional talking head with different audio emotion. We present emotion tag of \textbf{happy}, \textbf{sad}, and \textbf{angry} given  audio input of \textbf{\textit{fear}}. }
\label{fig:add_dv}
\end{figure*}

We provide Figure~\ref{fig:abla} to visualize the reconstruction results of different loss implementations. On both datasets, we observe that $ \mathcal{L}_{cord} $ improves the expression in the eye region, showing a brighter and more energetic expression  in the green box.  $ \mathcal{L}_{shape} $ further enhances the generation quality by displaying more accurate lip motion. By combining  $ \mathcal{L}_{cord} $ and $ \mathcal{L}_{shape} $, the results show less vagueness in the regions of the lips and teeth. 

We then report quantitative ablation studies of loss implementation in Table~\ref{tab:abla}. We adopt the root mean squared error (RMSE) for the expression parameters from the audio-expression module. The contrastive loss $\mathcal{L}^{const}_{reg}$ enhances the audio-expression module by alleviating the deterioration caused by person-specific audio features. The CLIP loss $\mathcal{L}_{cord}$ improves expression recognition accuracy, making the faces closer to the target emotion. Lastly, adding identification shapes with $\mathcal{L}_{shape}$  results in a lower AUE, indicating more vivid facial motion. 

We also report the hyperparameters of the loss functions, specifically the $\rho$ in the contrastive loss $\mathcal{L}^{const}_{reg}$ and weight set $ \{\lambda_{photo}, \lambda_{cord},\lambda_{shape}\}$ for the reconstruction loss $ \mathcal{L} $. In the third part of Table~\ref{tab:abla},  $\rho = 0.5$ performs the best, while $\rho = 1$ is is the worst across the two datasets. This indicates that when the weight of other individual components increases, it decreases the performance of expression parameter prediction. In the bottom part of Table~\ref{tab:abla}, we can obverse by up-weighting the $\mathcal{L}_{cord}$, the ER value is increasing,the ER value increases, which is consistent with the findings of the  $\mathcal{L}_{cord}$ ablation study in second part of Table~\ref{tab:abla}. However, When increasing  the $\{\lambda_{cord},\lambda_{shape}\}$ at, the same time, the reconstruction quality and emotion representation both decrease.  We eclectically apply $ \{\lambda_{photo}, \lambda_{cord},\lambda_{shape}\} = \{1,{1e-3}, {1e-9}\}$ for our experiments, as it provides the top two performances.

\begin{table*}[t]
\centering
\setlength{\tabcolsep}{2.1mm}
\caption{Ablation Studies of  loss implementation. We display the best outcome in bold and second-best in underline. \label{tab:abla}}{\begin{tabular}{@{}cccc|cccc|cccc@{}}
\toprule
\multirow{2}{*}{ $\mathcal{L}_{reg}$} & \multirow{2}{*}{$\mathcal{L}^{const}_{reg}$ } & \multirow{2}{*}{$\mathcal{L}_{cord}$} & \multirow{2}{*}{$\mathcal{L}_{shape}$ } & \multicolumn{4}{c|}{\textbf{MEAD}}                                    & \multicolumn{4}{c}{\textbf{CREMA-D}}                                  \\ 
                      &                        &                      &                      & \textbf{RMSE} $\downarrow$ & \textbf{M/F-LMD} $\downarrow$& \textbf{AUE} $\downarrow$& \textbf{ER} $\uparrow$& \textbf{RMSE}$\downarrow$ & \textbf{M/F-LMD}$\downarrow$ & \textbf{AUE}$\downarrow$ & \textbf{ER} $\uparrow$\\  \midrule
\ding{52}        &                &               &               & 0.45          & -              & -            & -  & 0.39          & -              & -            & -          \\
\ding{52}        & \ding{52}          &               &               & \textbf{0.27}          & -              & -            & -    & \textbf{0.16}          & -              & -            & -           \\ \midrule
\ding{52}        & \ding{52}          & \ding{52}         &               & -             & 1.176$/$1.099          & 1.52         & 66.53   & -             & 1.233$/$1.179          & 1.79         & 69.80       \\
\ding{52}        & \ding{52}          &               & \ding{52}         & -             & 1.104$/$1.051          & 1.73         & 63.71 & -             & 1.208$/$1.095          & 1.58         & 61.23       \\
\ding{52}        & \ding{52}          & \ding{52}         & \ding{52}         & -             & \textbf{1.031$/$0.981}          & \textbf{1.41}     & \textbf{73.28} & -             & \textbf{1.191$/$1.035}          & \textbf{1.55}         & \textbf{75.98}       \\ \midrule  
\multicolumn{4}{c|}{$\rho = 0.2$}                                 & 0.35             & -              & -            & -            & \underline{0.22}             & -              & -            & -     \\  
\multicolumn{4}{c|}{$\rho = 0.5$}                                &                \textbf{0.27}             & -          & -   & - & \textbf{0.16}              & -          & -       & -  \\  
\multicolumn{4}{c|}{$\rho = 0.7$}                                 & \underline{0.29}             & -              & -            & -            & 0.25             & -              & -            & -     \\  
\multicolumn{4}{c|}{$\rho = 1$}                                 & 0.31             & -              & -            & -            & 0.27            &               & -            & -     \\  \midrule
\multicolumn{4}{c|}{$ \{\lambda_{photo}, \lambda_{cord},\lambda_{shape}\} = \{1,{1e-2}, {1e-9}\}$}                      & -             & 1.158$/$1.094              & 1.56            & \textbf{74.33}            & -             & 1.271$/$1.096              & 1.61            & 75.01      \\ \multicolumn{4}{c|}{$ \{\lambda_{photo}, \lambda_{cord},\lambda_{shape}\} = \{1,{1e-3}, {1e-8}\}$}                      & -             & \textbf{1.004$/$0.853}              & 1.43            & 72.11            & -             & 1.215$/$1.050              & 1.56            & 71.45      \\  \multicolumn{4}{c|}{$ \{\lambda_{photo}, \lambda_{cord},\lambda_{shape}\} = \{1,{5e-3}, {5e-9}\}$}                      & -             & 1.085$/$1.098              & 1.49            & 72.48            & -             & \textbf{1.155$/$1.007}              & 1.70            & 73.69      \\ \multicolumn{4}{c|}{$ \{\lambda_{photo}, \lambda_{cord},\lambda_{shape}\} = \{1,{1e-2}, {1e-8}\}$}                      & -             & 1.091$/$1.072              & 1.55            & 69.79            & -             & 1.187$/$1.066              & 1.73            & 70.92      \\ \multicolumn{4}{c|}{$ \{\lambda_{photo}, \lambda_{cord},\lambda_{shape}\} = \{1,{1e-3}, {1e-9}\}$}    &                -             & \underline{1.034$/$0.981}          & \textbf{1.41}     & \underline{73.28} & -             & \underline{1.191$/$1.035}          & \textbf{1.55}         & \textbf{75.98}    \\ 
\bottomrule
\end{tabular}}
\vspace{-0.1in}
\end{table*}

\end{document}